# Archaeological Sites Detection with a Human-AI Collaboration Workflow

Luca Casini [a,1], Valentina Orrù [b], Andrea Montanucci [a], Nicolo Marchetti [b], Marco Roccetti [a]

[a] University of Bologna, Department of Computer Science and Engineering, Bologna, Italy;
[b] University of Bologna, Department of History and Cultures, Bologna, Italy

[1]To whom correspondence should be addressed. **E-mail:** luca.casini7unibo.it

**Abstract**
This paper illustrates the results obtained by using pre-trained semantic segmentation deep learning models for the detection of archaeological sites within the Mesopotamian floodplains environment. The models were fine-tuned using openly available satellite imagery and vector shapes coming from a large corpus of annotations (i.e., surveyed sites). A randomized test showed that the best model reaches a detection accuracy in the neighborhood of 80%. Integrating domain expertise was crucial to define how to build the dataset and how to evaluate the predictions, since defining if a proposed mask counts as a prediction is very subjective. Furthermore, even an inaccurate prediction can be useful when put into context and interpreted by a trained archaeologist. Coming from these considerations we close the paper with a vision for a Human-AI collaboration workflow. Starting with an annotated dataset that is refined by the human expert we obtain a model whose predictions can either be combined to create a heatmap, to be overlaid on satellite and/or aerial imagery, or alternatively can be vectorized to make further analysis in a GIS software easier and automatic. In turn, the archaeologists can analyze the predictions, organize their onsite surveys, and refine the dataset with new, corrected, annotation.

**Significance Statement**

In this paper we describe the use of a pre-trained neural network for semantic segmentation, fine-tuned on annotated images of archaeological sites from the Mesopotamian floodplain. Integrating human expertise, our models reached a detection accuracy of 80%. We also propose a workflow where archaeologists and AI are collaborating: the model highlights the presence of a site and its predictions can be stitched together to create a huge overlay of predictions or converted into vector shapes, which archaeologists can import into a GIS software, speeding up the remote sensing phase during ground survey preparation. After reviewing the predictions, the experts can in turn refine and extend the dataset, improving the model.

## Introduction

This paper documents the outcomes of a collaboration between data scientists and archaeologists with the goal of creating an artificial intelligence (AI) system capable of assisting in the task of detecting potential archaeological sites from aerial or, in our case, satellite imagery. This procedure falls into the domain of Remote Sensing (RS), which indicates the act of detecting and/or monitoring a point of interest from a distance. In the world of archaeology this operation has become invaluable with the availability of more and better imagery from satellites that can be combined with older sources of information (e.g., the CORONA satellite imagery) to spot a larger number of archaeological sites as well as tracking their successive degradation due to anthropic factors. Depending on the area of investigation and the size of the archaeological features being surveyed, the effort necessary, especially in terms of time, can be huge for the researcher.

This collaboration aimed at solving exactly this issue by using deep learning models to streamline, but not completely automate, the process. Thus, we set out to train a model on a dataset of geo-referenced shapes of all known sites scattered throughout the southern Mesopotamian floodplain (which represents a sufficiently coherent geo-morphological region). As the project went on, a number of issues emerged that make this problem particularly hard to tackle and lead to an important reflection on the use of deep learning in general and its relationship to human experts. The dataset, while may be considered a very large one for near eastern archaeology with its almost 5,000 sites, is hardly sufficient for training a model as large as the state-of-the-art ones we see in use today and, perhaps more significantly, contains many cases that are visible only on certain old imagery.

The first issue is commonly solved in machine learning by leveraging transfer learning and using pre-trained models that are then fine-tuned on the data at hand. The second one, however, puts both training and evaluation in jeopardy, as the model is pushed to make wrong classifications during training and even if it learned robust representations that ignore the bad examples, we would then have a hard time detecting what is a mistake by the model and what is a mistake in the labels.

We believe that the only way out of this conundrum is through a human-in-the-loop approach. For this reason, throughout the paper we highlight the importance of integrating domain expertise during the training and evaluation phase of our experiments, since that was crucial in improving the dataset used and, in turn, the model. The final outcome of this iterative process is a model capable of obtaining a detection accuracy of around 80%.

Based on these egregious results, we envision a tool for human-AI collaboration to support the archaeologists in the remote sensing operations (rather than replace them) and propose a new kind of workflow, enhancing both their task and the model by providing improved data after every use. All the results were achieved using open-source software and models, as well as openly available data (imagery, annotations) and computational resources (Google Colab), making this kind of work highly accessible and replicable even in resource-constrained research environments. All code and resources mentioned are available at https://bit.ly/PNAS_floodplains .

## Research Background

### The Mesopotamian Floodplain

The southern Mesopotamian floodplain is a crucial region for understanding the complex interplay between the spatial clustering of human communities and the development of irrigated farmland in an otherwise semi-arid environment [1]. Robert McCormick Adams' surveys in the area [2–4] were carried out according to standards that were unparalleled for the time: he used a set of aerial photographs from 1961 to locate potential sites and map canals whose traces were visible on the surface; he was systematic in recording sites ranging in time from the later 7th millennium BCE to the Ottoman period; above all, he was acutely aware of the historiographical potential of his survey work, which resulted in a powerful interpretation of settlement patterns and hydraulic activities [4].

After a long halt to fieldwork resulting from political instability, archaeological research resumed in southern Iraq in recent years (see [5] for an overview). In this area sites are usually referred with the Arabic word for mound, "Tell." The color and shape of these hills makes them especially visible from aerial and satellite imagery, which led to the use of remote sensing as a viable strategy to discover their location.

As Tony Wilkinson puts it *"Tells comprise multiple layers of building levels and accumulated wastes built up through time, in part because the locus of occupation has remained stationary. Tell settlements frequently are defined by an outer wall that both contained and constrained the accumulated materials, thereby restricting their spread [...]. The tell is by no means the sale locus of occupation [...]. Outer or lower towns [...] often appear as low humps or simply artifact scatters around tells, and they can extend the total occupied area of a site several fold"* [6].

In Mesopotamia, tells are often only slightly more elevated than the surrounding countryside, often being prone in such cases to artificial leveling in order to gain irrigable agricultural areas. Thus, the automatic detection of sites in such a dynamic environment is a highly complex operation, although contrasts are sufficiently marked to justify the attempt.

**Remote sensing**

By remote sensing one may refer to the use of any sensor (i.e., temperature, humidity, hyper-spectral, satellite images etc.) for detecting or monitoring a point of interest without the need of personally visiting it. This approach is relevant to a variety of fields, but solutions that work in one domain may not translate to others.

Locating archaeological sites remotely was certainly possible even before the advent of modern computer technology by using aerial photographs and topographical maps of the area to be investigated, but today it is easier to combine multiple sources, using sensors of different nature or from different points in time, to get a more complete picture of the environment, especially since it can be changing due to natural or anthropic factors [7–9]. Depending on the characteristics of the sites, certain representations can be helpful like elevation models obtained from stereoscopic images or the use of parts of the electromagnetic spectrum other than visible light like infrared or radio waves [10, 11].

LiDAR is also becoming popular as it gives extremely high-resolution images, but it can be difficult to employ as it requires to be mounted on some kind of airborne craft like drones [12]. The problem with these "unusual" types of sources is that they might not be available for every location or not have a high enough resolution for the task at hand. On the other hand, good quality color images of virtually any location on the planet is easily and freely available, largely due to the popularity of online services like Google Maps or Bing Maps.

**Deep Learning for Remote Sensing and Archaeology**

Deep learning has found multiple uses in every field of application and archaeology is no exception. It can help in classifying objects and text, finding similarities, building 3D models and, as this paper illustrates too, the detection of sites [13–17]. A difficulty in dealing with such a model is that it requires domain experts in both archaeology and deep learning to come together, but it may also depend on the amount of data available. Neural networks are notoriously data hungry, and archaeology is a "slow data" field as Bickler put it [18]. Nonetheless, there are a few recent examples of deep learning being successfully applied to site detection in a variety of different scenarios [19–22]. Most applications either use neural network to perform a classification task, with tiles sampled from maps that are marked as containing the site of interest or not, or as segmentation tasks where the individual pixels are classified, and the result is the prediction of a shape corresponding to the site. In this paper we will use the second approach, as described below.

**Semantic Segmentation**

Semantic segmentation is the task of dividing an image into parts that correspond to units with a specific meaning. These can correspond to a specific subject (e.g., the outline of persons, vehicles, etc.) or to a generic category that encompasses multiple entities (e.g., buildings, backgrounds, etc.). In the context of this paper, we only have two categories: one for mounded (tell) sites and another one for everything else. Segmentation can be performed with various techniques that perform pixel-level classification. A very common approach uses pre-computed features, extracted by some algorithm, or manually engineered, which are then classified by a Random Forest algorithm [23]. The current state of the art is represented by end-to-end systems based on deep learning with convolutional neural networks. For this approach, the introduction of U-Net by Ronnenberger in the context of medical imaging represented a milestone [24]. This work leverages a more recent architecture, called MA-Net [25], which can be thought of as an upgrade of the U-Net architecture with the attention mechanism. While it was developed in the context of medical imaging it has found use also in remote sensing tasks [26, 27]. In the methods section we will provide more details.

**Previous Work and Limitations**

In a previous paper we tried to tackle this same problem using an image classification approach where the map was divided into tiles [28]. In that experiment however the dataset was an order of magnitude smaller, and we had to resort to aggressive data augmentation in order to boost performance. The best model obtained an AUC score of around 0.70 but when tested on an unseen portion of map it showed its limits in that it predicted many false positives while also missing some sites. The biggest trade-off of this tile-based classification approach is between the size of the tiles and the granularity of the predictions with bigger squares that are more practical but result in a loss of detail. There is also the problem of dealing with sites that land on the edge of a tile. A solution we tried was creating a shingled dataset with in-between tiles to fill the gaps. This however greatly increased the amount of prediction to be created. Finally, most models for image classification are bound by the use of a fixed size of input which can be a huge limit when dealing with maps. In this new experiment, given the increased size of the dataset, we decided to leverage image segmentation models with fully convolutional layers which address both the limits in input size and the granularity trade-off.

## Materials and Methods

In this section we first describe the dataset used, which was built starting from openly available resources and then the open-source models we fine-tuned on that dataset.

**Vector shapes for archeological sites**
We started with a dataset of geo-referenced vector shapes corresponding to contours of known mound sites in the survey area of the Floodplains Project that spans 66,000 km^2, as shown in Figure 1. The dataset - developed at the University of Bologna by filing all published archaeological surveys in the area and geo-referencing anew the sites catalogued therein (https://floodplains.orientlab.net ) contains 4,934 shapes, thus all referring to sites which had been confirmed by ground truthing and by the associated study of the surface scatter of artifacts.

Since the dataset was compiled as a comprehensive source of information for archaeologists rather than specifically to train a machine learning model, we needed to filter out some examples that provided no information and could actually impair the learning process. We started by removing the top 200 sites by area as these were considerably bigger than the rest of the dataset and visual inspection confirmed that they follow the shape of areas that are not just simply mounds. The number 200 emerges from noticing that these sites have an area bigger than the square region we use as an input and could thus result in a completely full segmentation mask which would not be very helpful. After a discussion between data scientists and archaeologists we convened that this was a good heuristic solution.

Additionally, we filtered out 684 sites that either presented a very small area or were earmarked by the archaeologists as having been destroyed. In particular the size threshold was set at 0.1 degrees squared (roughly equal to 1,000 m^2). These very small sites actually correspond to a generic annotation for known sites with unknown size or precise location.

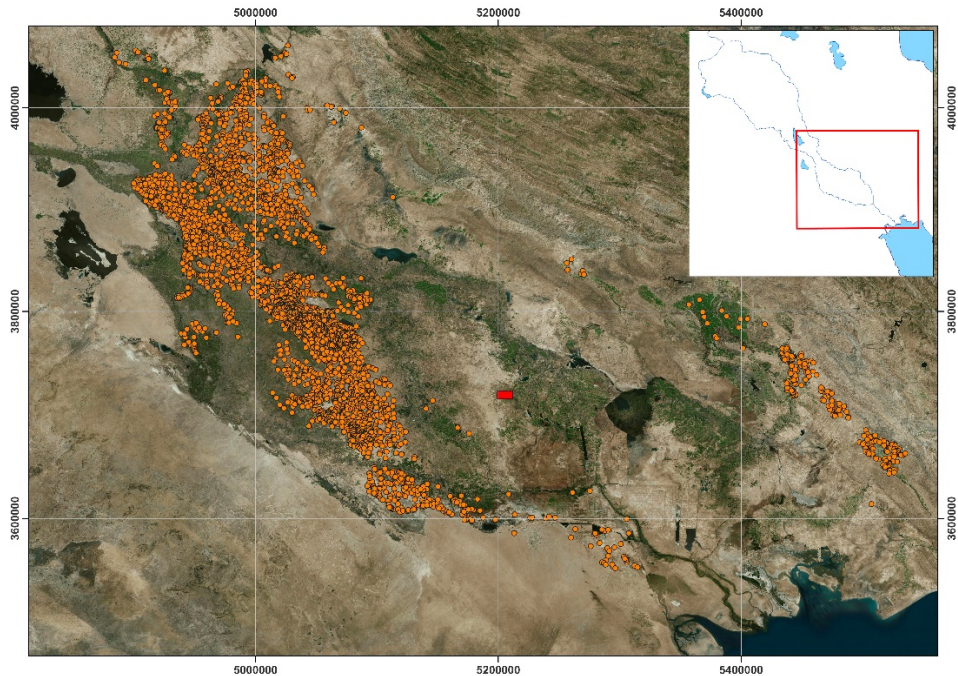

*Figure 1 Investigation area. Orange dots represent surveyed sites in the Mesopotamian floodplain. The red rectangle is a selected test area in Maysan.*

**Creating the input images**
To generate a dataset of images to fine-tune our pretrained model we imported the shapes mentioned above into QGIS (an open-source GIS software) [29] and using a Python script we saved a square of length L centered on the centroid of site which contains only satellite imagery from Bing Maps (we also considered Esri imagery but found that in this particular area they are the same). We then saved the same image without a basemap but with the site contours shown, represented as a shape filled with a solid color, to serve as our ground truth masks.

In the first experiments we set L to be 1,000 meters, but we imagined that increasing the size of the prediction area could be beneficial due to the inclusion of a larger context. Consequently, we also tried using L = 2,000 m and obtained improved performance overall.

From the starting square image, we randomly crop a square of length L/2 to be used as the input. This ensures that the model does not learn a biased representation for which sites always appear at the center of the input and additionally serves as data augmentation. Beside this crop, we also augment the dataset by applying a random rotation and mirroring, as well as a slight shift in brightness and contrast, all these operations being applied in a different manner at each training iteration. When extracting from QGIS, we saved images with a resolution of around 1 pixel per meter (1,024 pixels for 1,000 meters, double that for the model with increased input size) but the inputs were then scaled down to half of that to ease computational requirements while having low impact on the overall performance [30].

Finally, we introduced 1,155 images with empty masks (no sites to predict) sampled from locations suggested by the archaeologists. These include highly urbanized areas, intensive agricultural areas, locations subject to flooding (i.e., artificial lakes and basins) and rocky hills and mountains.

The number was chosen arbitrarily, taking into consideration the size of each suggested area and of the tiles. The final number of images is thus 5,025. We split the dataset into a 90% training set and a 10% holdout test set, stratifying the "empty" images we added.  10% of the training set was also randomly selected to be used as a validation set.

We tried integrating CORONA imagery as an additional input [31], as in the usual archaeological workflow that historical imagery is very useful (since it refers to a situation so much less affected by development) and often combine with the satellite base-maps and the topographical maps (but since CORONA were used as a complement, we do not pursue automatic detection on them only and thus sites destroyed after the 1970s have been excluded from the analysis). After importing the imagery into QGIS, we followed the same procedure to create the inputs, ensuring the crop operation was equal for both Bing and CORONA images.

**Semantic segmentation models**
This project started as an experiment to investigate the viability of pretrained semantic segmentation models as tools for detecting sites. For this reason, we decided to compare pretrained open-source models made available as part of a library written in PyTorch. The library allows one to choose an encoder convolutional neural network for feature extraction and a segmentation architecture independently, as well as providing a number of different loss functions [32].

In a previous preliminary paper, we experimented with different choices of architecture, encoders and loss functions [30]. We compared U-Net versus MA-net, Resnet18 versus Efficientnet-B3 and Dice Loss versus Focal Loss. The performance differences were small, within a few percentage points at best, which could be very well explained by fluctuations due to the random data augmentation.

Nonetheless, we took the best model which uses MA-net, Efficientnet-B3 and Focal Loss, trained for 20 epochs. We further tested for the effects of our filtering procedure (slightly improved from the previous work), and additionally experimented with the introduction of CORONA imagery and increased the input size.

**Tepa sites in Uzbekistan**
We also performed an additional test on another large dataset (https://www.orientlab.net/samark-land/) elaborated by the Uzbek-Italian Archaeological Project at Samarkand [33]. Given the similarity between the Tell in the Mesopotamian floodplain and the Uzbek *Tepa*, we wanted to see if the model was able to detect those sites without the need of additional retraining.

The dataset features 2318 point annotations categorized in different ways which also come with attributes related to their preservation states. We selected only sites classified as either *Tepa* or *Low Mound,* with the *Well-preserved* label. The final number of sites ends being 215: 148 Tepa and 67 Mounds. The actual test set images were created following the same procedure described above.

# Results

## Mesopotamia

First, we present the results in terms of average Intersection-over-Union (IoU) score on the test dataset. IoU does not directly relate to the performance in detecting the sites, but only represents the degree of correspondence between the predicted shape and annotation in the dataset. Still, it gives us an idea of how the model behaves and helps us select the best one. Table 1 summarizes the results for all models on the holdout dataset, as described in the Methods section.

Note that, for each model, we report a mean score and the associated standard deviation. This is due to the fact that we are performing a random crop on the images, even on the test set, and thus we run ten tests with different crops to average out this effect.

| Model | Input | Filter | Iou | St.Dev |
|---|---|---|---|---|
| Model 1 | Bing 1k | | 0.7417 | 0.0038 |
| Model 2 | Bing 1k | ✓ | 0.7810 | 0.0054 |
| Model 3 | Bing + CORONA 1k | | 0.7406 | 0.0039 |
| Model 4 | Bing 2k | | 0.7977 | 0.0034 |
| Model 5 | Bing 2k | ✓ | 0.8154 | 0.0035 |
| Model 6 | Bing + CORONA 2k | ✓ | 0.8345 | 0.0018 |

*Table 1 IoU scores for the different experimental setup we tested. The standard deviation comes from the repeated testing used to average out random cropping.*

The first thing that can be noticed is the marked improvement given by the increase in the input size. We imagine that the larger area provides more context to the predictions and makes the model more accurate.

As important is also the inclusion of the filtering procedure that results in a bump in performance regardless of the input size.

Finally, the use of CORONA imagery is a bit controversial. For the smaller input size, it seems to provide no benefits (the lower error score is within the margin of error) and we can hypothesize this is due to the low resolution of these photos. With larger areas they instead seem to provide an increase in performance, maybe again due to the larger context. Inspecting the prediction, however, revealed the absence of a marked difference, perhaps meaning the IoU is increasing just as the result of slightly more precise contours.

## Detection Accuracy

To further assess the results, we moved on to detection accuracy. First, we transformed the raster predictions from the model into vector shapes using the well-known library GDAL (Geospatial Data Abstraction Library) [34] and then we looked for the intersection between the site annotations and the predictions. To obtain smoother shapes, before the conversion we first applied a Gaussian blur to the prediction rasters and then clipped values above a certain threshold (0.5, but the number can be changed for a more or less sensitive model) to 1.0, while everything else would be set to 0.0.

This automatic evaluation gives good but not too exciting results, with an accuracy score of 0.6257 for Model 5 and 0.6008 for Model 6. A model able to find two out of three sites would already provide a good starting point for human analysis. However, archaeologists must provide a verification of the predictions and differentiate the cases in which the model commits proper mistakes from those in which it makes justifiable errors that a human would do too [35].

First of all, there are a considerable number of sites that are no longer visible from present day satellite images and were not filtered from the dataset. This was expected as only 50% of the annotations had additional information and even less contained indication of their visibility. Those sites should not be considered as False Negatives but rather as True Negatives.

When it comes to predictions marked as False Positive, sometimes the model predicts another site close by, instead of not the one being tested. This can be considered a mistake or not depending on the nature of the ``missed'' site. In one case we have a site that is no longer visible, so the prediction is actually a True Positive. On the other hand, it can be a site that is still visible but maybe less so than another one in the picture. In this situation we could either consider both a false negative and a true positive, or just as a true positive given that, in a real world scenario, the closeness to other sites would result in a useful suggestion as the human expert, who would then be able to retrieve them all. Alternatively, we could avoid considering non-visible sites altogether, but the difference would be minimal with accuracy 0.7837 and recall 0.8201.

Lastly, some predictions were actually present in the outputs but too faint for the cutoff threshold we imposed. We did not adjust for those errors, but they indicate a possible approach for interaction: using predictions as overlays and manually looking at the map. Alternatively setting a lower threshold could solve the problem.

Table 2 summarizes the results for the automatic evaluation and the adjusted values after the human evaluation highlighted non-visible sites. The adjustment raises accuracy and recall to around 80, giving a more objective idea of the actual model performance.

| Model | Evaluation | TP | TN | FP | FN | Accuracy | Recall |
|---|---|---|---|---|---|---|---|
| Model 5 | Automatic | 228 | 98 | 70 | 125 | 0.6257 | 0.6459 |
| Model 5 | Adjusted | 258 | 185 | 40 | 68 | 0.8040 | 0.7914 |
| Model 6 | Automatic | 209 | 104 | 57 | 151 | 0.6008 | 0.5806 |
| Model 6 | Adjusted | 239 | 197 | 27 | 88 | 0.7913 | 0.7309 |

*Table 2  Site detection performance for the best models. Automatic evaluation considers the labels as they come, adjusted evaluation compensates for incorrect labels with a human in the loop.*

It is interesting to see how Model 6, which got a higher IoU score, seems to actually be performing worse now. Looking at the images, it appears that this model is a little bit more restrained and cautious, resulting in less positive predictions and thus less False Positives. In turn, this can result in a higher IoU because it reduces the Union term, and, if areas are a little bit more precise, it even raises the Intersection term. However, for detection's sake, we need the presence of an intersection rather than a perfect match and in this situation the lower number of positives is punishing. Overall, the difference in accuracy is not excessive, so both models are useful and could be used in parallel, but we must also consider the additional complexity and cost of using two sets of input images which make Model 6 a bit cumbersome. For this reason, we moved on using just Model 5.

We concluded this subsection with Figure 2, which contains a few examples from the test dataset to display the quality of the model's outputs. Note how the colors correspond to probability values, and that faint areas would be cut off by the 0.5 threshold we use in creating the vector shapes. The model is very accurate at tracing the site outlines and in some cases (i.e., the first row in Figure 2) these are even more accurate than the ground truth with respect to current satellite imagery.

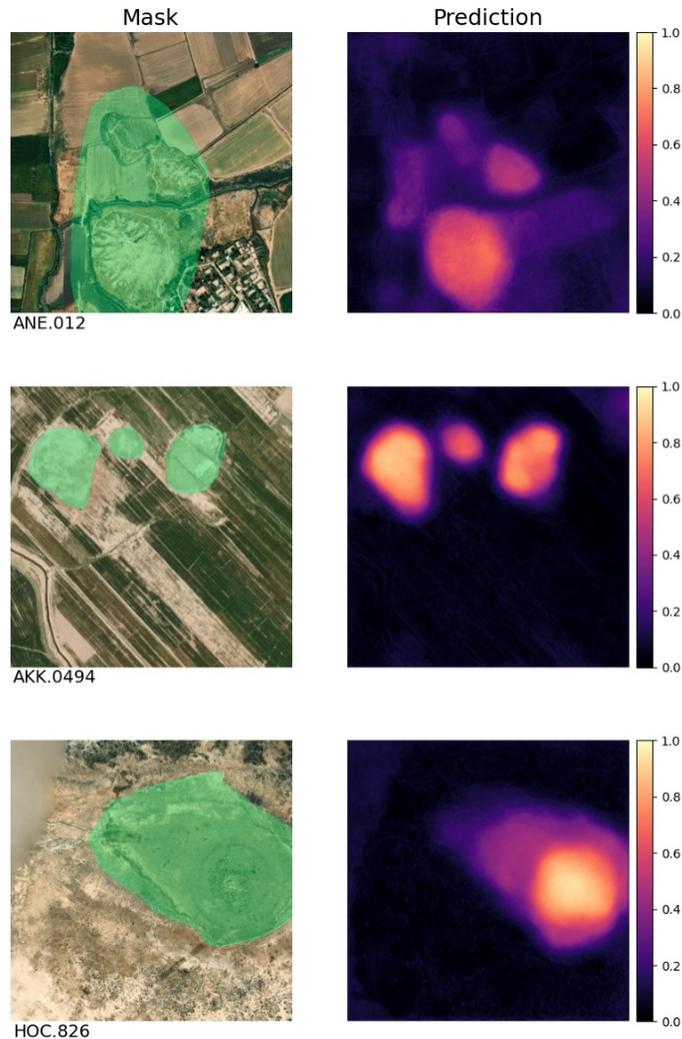

*Figure 2 A few sample predictions from the test set. On the left is the target mask overlaid on the input image. On the right the model output. The colorbar corresponds to classification probability. Note how the model is capable of matching accurately the site outline*

**A test in the Maysan province**
After assessing detection performance, we wanted to try the model on a rectangular area within the unsurveyed Maysan province for which we carried out remote sensing. This test had the goal of evaluating how many false positives the model would predict and to give an example of the mistakes the model makes in an operational scenario.

The area we selected contains 20 alleged sites and spans 104 km^2. Figure 3 shows the area with the annotation from the archaeologist and the prediction from the model. As it can be seen the model is able to recover 17 of the 20 sites while also suggesting around 20 more shapes (or less, depending on what is considered a single instance). Most of those suggestions are not useful but are also easily and quickly sifted out by an expert eye, especially in context, given their size or their location.

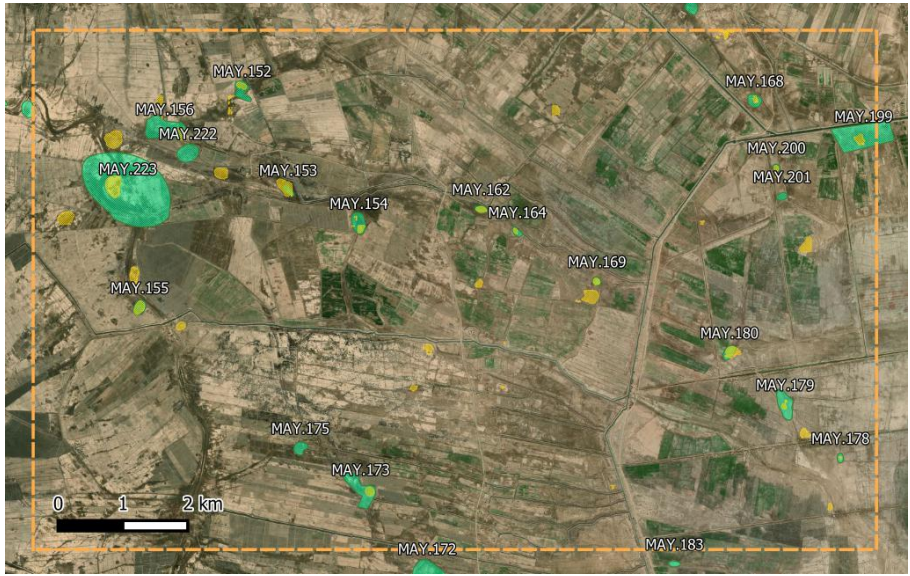

*Figure 3 Maysan province test area (orange) with sites remotely identified by archaeologists (turquoise) and model predictions (yellow). The sites identified by the trained eye and the model are equivalent and, most importantly, the model is able to ignore areas without significant features.*

Figure 4 instead shows an overlay produced by stitching together the various predictions and using the probabilities values as a sort of heatmap. "Hotter" colors correspond to higher probabilities while black indicates the absence of a site. The transparency is obtained through the use of the Overlay filter in QGIS.

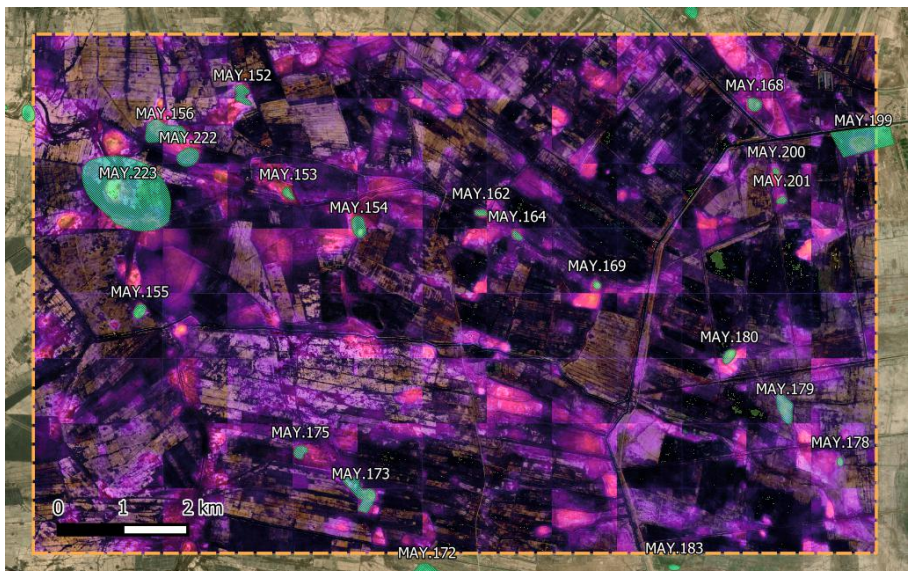

*Figure 4 Maysan test area prediction probabilities overlaid on top inside QGIS. This visualization allows the user to decide where to look instead of relying on a predefined threshold value.*

**Uzbekistan**

Unfortunately, human evaluation of the outputs showed that the model is able to correctly identify only around 25% to 30% of the sites in this region, depending on how thresholds are chosen. The remaining part contains either sites that are missed completely or sites that are somehow hinted either too faintly or inside a huge area that appears meaningless.

The reason for this severe drop in performance is most probably due to the different nature of the landscape in the region which in some locations appear to be way more urbanized and in general features more vegetation: thus, not all floodplain environments are similar enough for a direct cross-comparison. Furthermore, the conventions which lie behind the annotations in the Uzbek dataset might not be perfectly aligned with the Mesopotamian one further complicating the situation.

The only way of dealing with this problem here is probably to create a small dataset of selected Tepa sites and perform an additional round of transfer learning so that the model may grasp the new context and characteristics in the region.

**Discussion**

The results obtained can be considered satisfactory even if the IoU metric, when compared to other semantic segmentation applications, is not extremely high. When testing for detection performance, however, we found that the model is still able to detect most sites in the dataset, leaving us with good expectations for its use in other parts of the survey area. As the Uzbek test shows however, when it comes to new areas with similar sites but in a different context, performance may drop severely and a retraining phase, even with a smaller dataset, would be necessary. Future work may explore this research direction.

It is important to notice how evaluation metrics in this task seem to hit a wall when confronted with the fact that they are computed against annotations that oftentimes are not homogeneous and contain various spurious labels. In our case we coped with the fact that there are many sites that are only visible on some historical photographs or maps that are part of the dataset even if they do not provide useful examples. Fortunately, the model seems to be robust enough to learn useful concepts and ignore these confounding data points. Still a smaller, cleaner dataset could drastically improve performance while also reducing computational load. Obviously, such cleaning operations would be a massive investment in terms of time and archaeologists would rather spend it actively searching for sites themselves, instead.

Our model, however, opens up the possibility of going through already surveyed areas automatically and then producing a list of predictions that contrast the annotations to be manually reviewed. Subsequently a new, cleaner dataset could be assembled by the archaeologists and a new improved model could be trained. This same procedure also works in applications to new areas, where novel predictions can be manually checked and added to a new dataset overtime.

In addition to the automatic procedure, the model could also be used to produce an overlay to guide the eye of the archaeologist inside a GIS software. This graphical approach allows the users to also compare the overlay with other maps they might be using and use their expertise to infer the existence of a site based on all contextual information they have. We only tried this approach on a small area as shown in Figure 4 but the computation could be easily scaled up to cover huge areas, as it takes less than a second to produce an output and there is no need to complete the operation in one go anyway. The only shortcoming of this method is the evident ridge between different input images. In theory, semantic segmentation could work with inputs of arbitrary size, but doing so requires a huge amount of memory which might not be available. A solution might be the creation of overlapping prediction maps that would then be averaged, trading off computational time for increased precision.

Figure 5 summarizes the use we envision for the model we described. Starting from the dataset the model produces prediction masks that we can manipulate through post-processing to obtain either a vector shapefile that can be used for automatic evaluation and detection of sites. At this stage the user has the possibility of choosing a threshold to cut prediction off and the use of techniques to smooth the output shapes, like blurring or buffering the vectors. Similarly, the map overlay can be adjusted by selecting different graphical representations directly into the GIS software. The goal in this case is that of spotting sites that might go undetected by the automatic comparison because their probability is lower than the threshold, while still being distinguishable for a human. Each time the model is used, in either way, after reviewing the outputs the users would be able to obtain either a new set of annotations or a list of sites to be removed or relabeled.

If such a workflow is used by more than one team it could also greatly speed up the search efforts: the use of open technologies in this case makes the results easier to share between research groups, which could greatly help archaeology as a field [36].

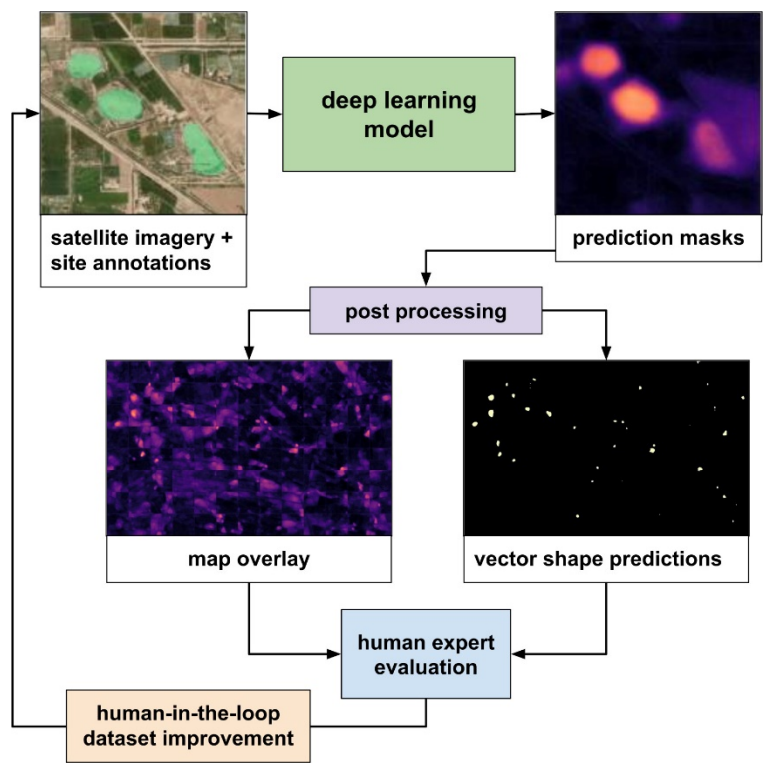

*Figure 5 A human-in-the-loop workflow based on our model. A model is trained from annotated images and provides predictions masks. The masks can be used as an overlay or vectorized. Human evaluation is conducted on the outputs and in turn a refined dataset can be created to improve the model.*

The experiments with CORONA imagery also hint at the possibility of combining more models, perhaps trained with different basemaps or a combination of them, and compare the prediction given by all of these. Especially if historical photos are present, we could end up with a dataset that also contains temporal information about when a site is visible and when it becomes undetectable. Use of stereoscopic images for the creation of elevation models could also benefit the task, if the resolution is sufficient to highlight the low mounds we are looking for.

**Conclusions**

We presented a deep learning model for detection of mounded archaeological sites in the Mesopotamian floodplain. The model was implemented using pretrained models for semantic segmentation, fine-tuned on satellite images and masks of the site shapes coming from a dataset containing almost 5,000 examples.

The result of our experiments is a model which obtains an IoU score of 0.8154 on the test dataset and detects sites with 80% of accuracy. This statistic accuracy however is adjusted for the considerable number of sites that appear mislabeled as they are no longer visible on modern satellite imagery. While we cleaned up the dataset to the best of our ability, many undetectable sites still remained. The model seems to be quite robust, however.

Following this result, we propose a workflow for the archaeologists to adopt, in which their already established remote sensing practices are supported and enhanced by the use of a model like our own. The outputs can be used both for very fast automatic detection, being aware of the mistakes this could introduce, or combined to generate a graphical overlay to direct the user's attention towards certain areas. In turn, the use of the model will result in new shapefiles and annotations that can be used for retraining and improving the model, as well as enabling further analyses.

**Acknowledgments**

FloodPlains Project. https://floodplains.orientlab.net/. The FloodPlains Project has been developed in the framework of the European Union project "EDUU – Education and Cultural Heritage Enhancement for Social Cohesion in Iraq" (EuropeAid CSOLA/2016/382-631), www.eduu.unibo.it , coordinated by Nicolò Marchetti.
The ongoing project "KALAM. Analysis, protection and development of archaeological landscapes in Iraq and Uzbekistan through ICTs and community-based approaches," funded by the Volkswagen Foundation and coordinated by N. Marchetti, www.kalam.unibo.it , has allowed a review of our data input and the development of the research presented in this paper. The CRANE 2.0 project of the University of Toronto provided the geospatial servers on which FloodPlains is running.